\PassOptionsToPackage{table,dvipsnames}{xcolor}
\documentclass[10pt,twocolumn,letterpaper]{article}

\usepackage[pagenumbers]{iccv} 

\usepackage{multicol}
\usepackage{multirow}
\usepackage{graphicx}
\usepackage{booktabs}
\usepackage[ruled, linesnumbered]{algorithm2e}
\usepackage{ulem}
\usepackage{caption}
\usepackage{diagbox}
\usepackage{makecell}
\usepackage{colortbl}
\usepackage{tablefootnote}
\usepackage{xparse}
\usepackage{soul}
\usepackage[table]{xcolor}
\usepackage{booktabs}
\usepackage{hhline}
\usepackage{verbatim}

\definecolor{iccvblue}{rgb}{0.21,0.49,0.74}
\usepackage[pagebackref,breaklinks,colorlinks,allcolors=iccvblue]{hyperref}

\def\confName{ICCV}
\def\confYear{2025}

\title{MMAIF: Multi-task and Multi-degradation All-in-One for Image Fusion with Language Guidance}

\author{
Zihan Cao\\
UESTC\\
{\tt iamzihan666@gmail.com}
\and
Yu Zhong\\
UESTC\\
{\tt yuuzhong1011@gmail.com}
\and
Ziqi Wang\\
UESTC\\
{\tt ziqiwang327@foxmail.com}
\and
Liang-Jian Deng\\
UESTC\\
{\tt liangjian.deng@uestc.edu.cn}
}

\begin{document}
\maketitle


\begin{abstract}
Image fusion, a fundamental low-level vision task, aims to integrate multiple image sequences into a single output while preserving as much information as possible from the input. However, existing methods face several significant limitations: 1) requiring task- or dataset-specific models; 2) neglecting real-world image degradations (\textit{e.g.}, noise), which causes failure when processing degraded inputs; 3) operating in pixel space, where attention mechanisms are computationally expensive; and 4) lacking user interaction capabilities.
To address these challenges, we propose a unified framework for multi-task, multi-degradation, and language-guided image fusion.
Our framework includes two key components: 1) a practical degradation pipeline that simulates real-world image degradations and generates interactive prompts to guide the model; 2) an all-in-one Diffusion Transformer (DiT) operating in latent space, which fuses a clean image conditioned on both the degraded inputs and the generated prompts. Furthermore, we introduce principled modifications to the original DiT architecture to better suit the fusion task. Based on this framework, we develop two versions of the model: Regression-based and Flow Matching-based variants.
Extensive qualitative and quantitative experiments demonstrate that our approach effectively addresses the aforementioned limitations and outperforms previous restoration+fusion and all-in-one pipelines. Codes are available at \url{https://github.com/294coder/MMAIF}.
\end{abstract}

\section{Introduction}
Image fusion aims to integrate image sequences from multiple modalities or a single modality but with varying camera parameters into a single image, while preserving the representative objects or information from the input sequences. For example, in multi-modal image fusion, such as near-infrared and visible image fusion (VIF)~\cite{ma2019infrared}, the goal is to highlight human or animal targets from the near-infrared data while retaining the color and texture information from the visible light image. In single-modal image fusion, such as multi-exposure (MEF)~\cite{zhang2021benchmarking} or multi-focus image fusion (MFF)~\cite{liu2020multi}, the aim is to preserve information from different parameters (\textit{e.g.}, exposure, focus). With the development of convolutional neural networks \cite{xu2020u2fusion,Zhao_2023_CVPR}, diffusion models~\cite{cao2024diffusion,zhong2025ssdiff}, Transformers~\cite{zhao2024image_film,dengtgrs2023}, and linear-RNN networks like Mamba~\cite{cao2024novel}, a large number of works are pushing the limits of the image fusion task.

\begin{figure}
    \centering
    \includegraphics[width=\linewidth]{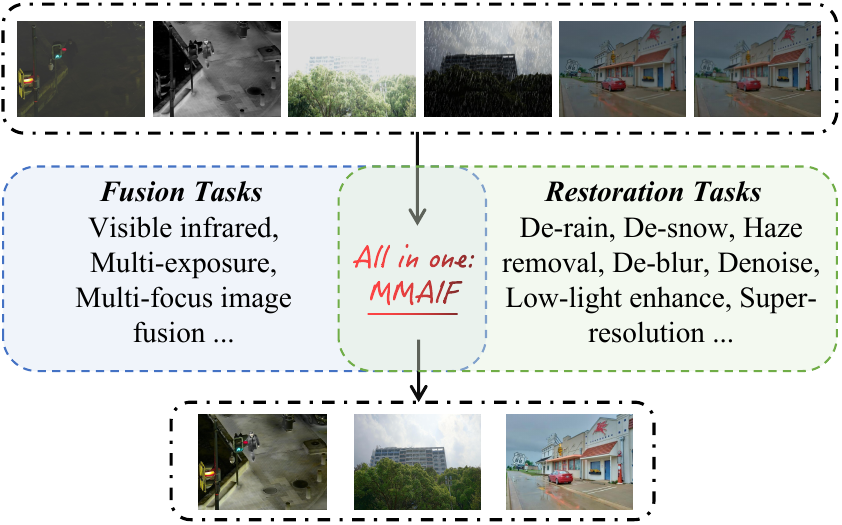}
    \caption{Main conception of the proposed MMAIF.}
    \label{fig:teaser}
    \vspace{-1.5em}
\end{figure}
However, previous works~\cite{zhang2023visible, tang2023datfuse, qu2022transmef, hu2023zmff} often overlook the fact that \textbf{\textit{real-world images may encounter various types of degradations, or even composite degradations}}, during the imaging process. In such cases, fusion networks trained on clean images often fail. A common solution is to \textbf{\textit{add one or more image restoration networks or algorithms before the fusion network}} (termed as restoration+fusion pipeline) to restore the images and then fuse the restored image sequences. This often introduces additional training or inference overhead and complicates the inference process. Furthermore, \textbf{\textit{multi-task image fusion requires multiple networks to be trained specifically for each task}}. For example, a network trained for VIF task cannot be directly adapted to MEF task, and a network trained for denoising cannot handle deraining scenarios. Therefore, there is an urgent need for a framework that can be trained and inferred under a multi-task, multi-degradation scenario. On the other hand, \textbf{\textit{the vast majority of fusion networks typically operate in pixel space}}, especially when the fusion network is Transformer-based. The quadratic complexity of the attention mechanism makes training costs and inference speeds even more unacceptable. Finally, image restoration tasks may require human guidance for better recovery, which is still under-explored. Note that recent works TextIF~\cite{yi2024textif} considered degradation in images and performs pixel space restoration and fusion, but it cannot adapt to multiple fusion tasks. DRMF \cite{tang2024drmf} and Text-DiFuse~\cite{zhangtextDiFuse} introduce two diffusion path integrations to restoration and fusion but make the training and inference complicated.

To address these limitations, we propose a multi-task, multi-image degradation, and language-guided image fusion framework, comprising two key components:
1) We design a practical degradation pipeline to simulate real-world image degradations and develop interactive prompts based on these degradations to guide the model. For example, the simulated degradation includes: noise, blur, and different weather effects.
2) We propose an all-in-one Diffusion Transformer (DiT)-based network operating in latent space to perform image restoration and fusion simultaneously, conditioned on the degraded images and prompts. Furthermore, we make principled and partial modifications to the original DiT architecture to better adapt it to the task, including 2D rotary position embedding (RoPE), attention value residual, bias inductive convolution before attention, and mixture-of-experts (MoE) gated linear units.
Building upon this, we conduct two version models that share a similar DiT backbone, including a deterministic regression-based and a generative flow matching-based model~\cite{lipman2022flow} to fulfill the learning process.

We summarize the proposed method and contributions as follows:
\begin{itemize}
    \item We propose a multi-task, multi-degradation all-in-one image fusion framework that unifies image fusion and restoration tasks into a \textit{single} model. This significantly alleviates the need to train multiple models and simplifies the inference procedure;
    \item Within the fusion framework, we introduce: a real-world image degradation pipeline and a modernized DiT architecture operating in the latent space to support multi-source degraded images and interactive prompt fusion. Two version models (\textit{i.e.}, regression and flow matching) are proposed to balance the latency and performance;
    \item Our fusion framework outperforms existing restoration + fusion methods and recent all-in-one image fusion methods on various degraded image fusion tasks, while also demonstrating superior inference efficiency.
\end{itemize}

\section{Related Works}
In the following sections, we will discuss the related works, focusing on the key elements of our proposed framework.

\subsection{Single and Multi-task Image Fusion Methods}
Previous image fusion methods \cite{ma2022swinfusion, zhao2023ddfm, zhu2024task, wang2024general, cao2024ttd, fc-former} have focused on network design, utilizing better loss functions, incorporating pre-trained prior information, and jointly training fusion and downstream tasks by leveraging downstream prediction. 
Some recent methods have recognized the complexity of training a separate network for each image fusion task. For example, U2Fusion~\cite{xu2020u2fusion} first utilizes a sequential task-learning approach to train an all-in-one model. PSLPT~\cite{wang2024general} employs a pseudo-Siamese Laplacian pyramid Transformer to differentiate different frequencies for multi-task learning.

\subsection{Degraded Image Restoration and Fusion}
Traditional image fusion models are typically trained on clean data pairs, but real-world images often contain degradations like noise, haze, or modality-specific artifacts. Fusion networks trained on clean images struggle with degraded inputs, as the goal is to produce a clean, well-fused output. A simple solution is to use separate restoration networks before fusion, but this complicates the pipeline and lacks automatic degradation-based model selection (Fig.~\ref{fig: main-framework}(a)). Additionally, restored images may fall out of distribution, causing fusion failures.
Recently, TextIF~\cite{yi2024textif} utilizes constructed degradation/clean/prompt pairs to automatically prompt the model to perform restoration (implicitly in the model) and fusion. Similarly, DRMF~\cite{tang2024drmf} and Text-DiFuse~\cite{zhangtextDiFuse} leverage multiple restoration diffusion models trained in pixel space, combined through a simple diffusion composition module to achieve fusion with restoration capabilities.

\subsection{Image Tokenizers and Generative Methods}
Image tokenizers \cite{flux2024, zhu2023designing, agarwal2025cosmos, zhao2024image} have become key components in image generation and multi-modal models. Formally, an image tokenizer consists of an encoder $E(\cdot)$ and a decoder $D(\cdot)$. The encoder maps an image $X\in \mathbb R^{H\times W\times c}$ to a set of latents $E(X)=Z\in \mathbb R^{h\times w\times z}$, where $H=h\times f$ (\textit{e.g.}, $f=8$ and $z=16$). The latent represents the original image in a more compact latent space and reduces the input size for subsequent models. The decoder remaps the latent back to an image $\tilde X=D(Z)$ in pixel space, aiming for the best reconstruction for the input $X$. 

\begin{figure*}[t]
    \centering
    \includegraphics[width=\linewidth]{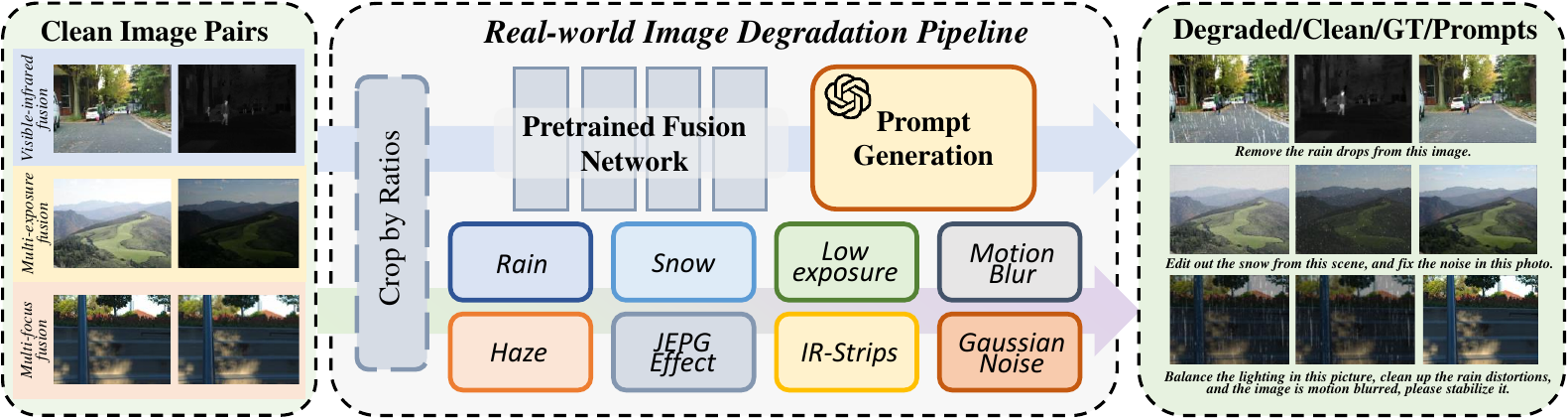}
    \caption{\textbf{The proposed real-world image degradation pipeline}. Different clean image pairs from multiple tasks are sent to a pre-trained fusion network to obtain the GT. The pairs are also fed into many compositional degradation operators (\textit{e.g.}, haze, rain) to get degraded pairs. While leveraging ChatGPT to generate restoration prompts, we can collect degraded/clean/GT/prompts pairs.}
    \label{fig:degradation_pipeline}
    \vspace{-1em}
\end{figure*}

Latent diffusion models~\cite{rombach2022high} and latent flow matching~\cite{dao2023flow} perform generative modeling in the latent space to generate images or videos. Taking flow matching as an example, consider a prior distribution $Z_0\in p_0$ and a generative distribution $Z_1\in p_1$. Flow matching models the generative process as an ordinary differential equation (ODE):
\begin{equation}
    dZ_t=v_\theta(Z_t,t)dt,
\end{equation}
where $v_\theta$ is called the velocity model. $v_\theta$ is trained by the flow loss:
\begin{equation}
    \mathcal{L}_{flow}=\mathbb E_{Z_1\in p_1, Z_0\in p_0}[\|v_\theta(Z_t,t) - (Z_1-Z_0)\|^2], \label{eq: flow-loss}
\end{equation}
where $Z_t=tZ_0+(1-t)Z_1$ and $t\in\mathcal U(0,1)$.
Generally, $p_0$ is chosen as the standard Gaussian distribution, and $p_1$ is the data distribution.

\section{Methodology}

In this section, we will describe the key components of the proposed framework: 1) the real-world image degradation pipeline involved; 2) the latent tokenizer used; and 3) how the original DiT architecture is adapted for multi-degraded-image sequence and prompt input, a series of modernization improvements at the architectural level, and how to train fast deterministic regression model and the generative flow matching model.

\subsection{Real-world Image Degradation Pipeline}
\label{sect: degraded-pipeline}
\begin{figure*}[t]
    \centering
    \includegraphics[width=\linewidth]{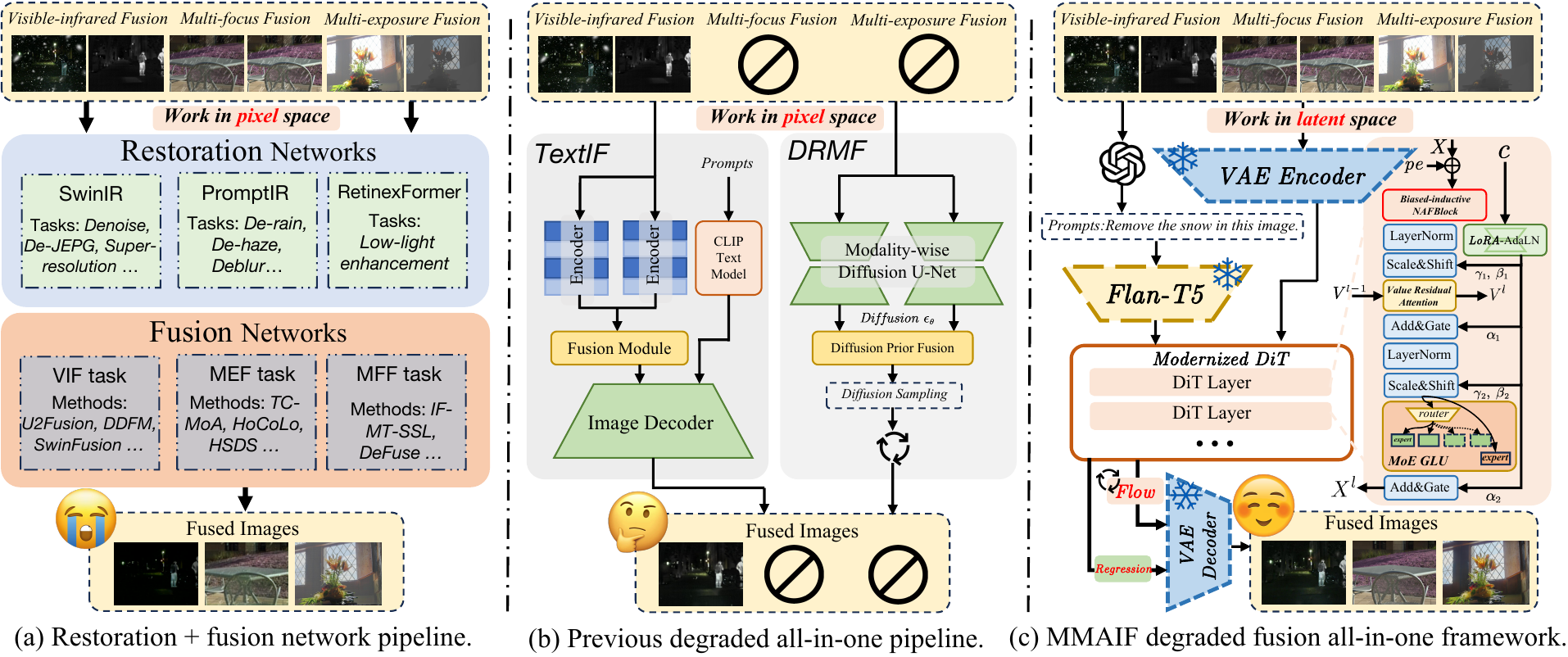}
    \caption{\textbf{Comparisons with previous pipelines}. (a) Naive restoration + fusion pipeline. It causes a complex inference process which restores to tiling the high-resolution images and needs to handle different degradation or fusion tasks by distinct models. (b) Recent all-in-one models only take account of multiple degradations but neglect task-level all-in-one and have large FLOPs when operating in pixel space. (c) Our framework operates in the \textit{latent space}, leveraging a modernized DiT. This enables the training of a unified \textit{task-level and degradation-level all-in-one model} that supports either fast regression or refined flow matching.}
    \label{fig: main-framework}
    \vspace{-1.3em}
\end{figure*}

To ensure the pre-trained DiT model adapts to real-world degradations as effectively as possible, we design a principled, task-specific image degradation pipeline, illustrated in Fig.~\ref{fig:degradation_pipeline}. We fix the scope of the image fusion tasks to VIF, MEF, and MFF tasks. For these three tasks, common degradation strategies include: Gaussian blur, motion blur, downsample, Gaussian noise, rain, haze, and snow. We additionally apply the following degradations: low exposure, low contrast, and IR dark strips for VIF task; and low contrast for MEF task; low exposure and high exposure for MFF task. When inputting each image pair into the pipeline, we uniformly sample $n$ types of degradations from the pipeline with a certain probability and a random ratio to crop (\textit{e.g.}, 16:9, 4:3, and 1:1) and apply them to the image pair.

It is worth noting that real-world degradations may not be limited to a single type (\textit{e.g.}, sleet). We set $n\in\{1,3\}$ to simulate these situations. Additionally, since VIF involves multi-modal sequences, applying weather effects such as rain requires different treatments for different modalities. Specifically, we uniformly apply weather degradation to the visible image and apply a certain degree of blur to the corresponding infrared image to better approximate the real world. To align with real-world degradation, for the haze degradation, we first use DepthAnything \cite{depth_anything_v2} to estimate the depth of the image and then apply the atmospheric light scattering formula \cite{zhang2017hazerd} to add haze, which is more realistic than traditional center-based haze addition.

To enable the training process of flow matching, we use pre-trained SwinFusion \cite{ma2022swinfusion} and DeFuse \cite{Liang2022ECCV} to collect the ground truth (GT) for each clean pair. When in the degradation stage, we do not apply any degradation on GT but only resize or clip into the same size as the degraded pairs. 
To provide the model with interactive capabilities, we use ChatGPT \cite{chatgpt} to generate 10 to 20 prompts for each type of degradation (\textit{e.g.}, ``please remove the noise of the image pairs''). These prompts are encoded and then input into the network during training to guide image restoration and fusion.
Finally, we collect all degraded/clean/GT/prompt pairs into a unified dataset for subsequent training. 
Recently, the dataset proposed in TextIF~\cite{yi2024textif} also features a degradation/clean/prompt structure. Differences with their dataset are discussed in the Suppl. Sect. {\color{iccvblue} 4}.

\subsection{Choices of the Image Tokenizer}

Since our framework operates in the latent space, the performance of the fused image is limited by the reconstruction performance of the tokenizer's decoder. Therefore, selecting a tokenizer with sufficiently good reconstruction performance is crucial. Furthermore, since both regression and flow matching operate in continuous space, quantization-based tokenizers \cite{esser2021taming} are not suitable. Consequently, we select Flux KL-VAE \cite{flux2024}, Asymmetrical KL-VAE \cite{zhu2023designing}, and Cosmos VAE \cite{agarwal2025cosmos} to compare. They are all set to $f=8$ and $z=16$. As shown in Tab.~\ref{tab:vae-types}, Cosmos VAE exhibits the best reconstruction performance; therefore, our primary experiments are implemented using this VAE.

\begin{table}[t]
    \centering
    \caption{\textbf{Different image tokenizer choices}. Cosmos VAE performs the best reconstruction. All metrics are tested on the average of SICE and LLVIP datasets.}
    \vspace{-0.5em}
    \label{tab:vae-types}
    \resizebox{\linewidth}{!}{
    \begin{tabular}{c|ccc}
    \toprule
       Tokenizers  & Flux KL-VAE \cite{flux2024} & Asy. KL-VAE \cite{zhu2023designing} & Cosmos VAE \cite{agarwal2025cosmos} \\
    \hline
        PSNR & 33.41 & 33.10 & 34.02 \\
        SSIM & 0.9227 & 0.9201 & 0.9367 \\
    \bottomrule
    \end{tabular}}
    \vspace{-1em}
\end{table}

\subsection{Modernized DiT for Image Fusion}
\newcommand{\newsubsec}[1]{\noindent \textbf{#1}}
In this subsection, we first introduce the preliminary of DiT and then propose a fusion adapted modernized DiT for all-in-one degraded image fusion.

\subsubsection{Preliminary of DiT} 
Original DiT \cite{rombach2022high} uses a Transformer as the backbone, and first patchify the input 2D latents shaped as $h\times w\times z$ with a patch size $p=2$, transforming them into 1D tokens shaped as $\frac{h\cdot w}{p^2} \times p^2z$. These tokens are then added with sine positional embeddings (PE) \cite{dosovitskiy2020image} to provide location information. The timestep is also encoded using sine PE, and the class label is embedded. These tokens are input into a series of Transformer blocks for processing. Each block consists of an attention block and a feed-forward network (FFN). 
To incorporate conditional information into the model, DiT uses adaptive layer norm (adaLN) with zero initialization:
\begin{equation}
    \text{AdaLN}(X,\beta,\gamma)=\gamma\cdot\text{LN}(X) + \beta,
\end{equation}
where LN is the layer norm, and $c$ denotes conditions which is the addition of class label and timestep embeddings in DiT.

Every Transformer block can be formulated as:
\begin{equation}
\begin{cases}
    \alpha_1,\beta_1,\gamma_1,\alpha_2,\beta_2,\gamma_2=\text{MLP}(c),\\
    X_{attn}=X+\alpha_{1}\cdot Attn(\text{AdaLN}(X, \beta_1,\gamma_1)), \\
    X_{FFN}=X_{attn}+\alpha_{2}\cdot FFN(\text{AdaLN}(X,\beta_2,\gamma_2)).  
\end{cases}\label{eq: ada-ln}
\end{equation}
FFN is implemented as a three-layer MLP.
After passing through several blocks, the features are output through the final AdaLN layer, unpatched back into 2D latent, and finally mapped back to pixels by the tokenizer decoder.

\subsubsection{Fusion Adapted and Modernized All-in-one DiT} 
\label{sect: modernized-dit}

The original DiT needs to be modified for the degraded image fusion task. 
We introduce some important modifications as listed below. The whole modified DiT architecture is shown in Fig.~\ref{fig: main-framework} (c). We ablate these modifications in Tab.~\ref{tab: ablations} to verify their effectiveness.

    \textbf{\textit{Mixture-of-experts (MoE)}}: Due to the need for simultaneous training across multiple tasks and degradations, the model capacity requires a certain degree of expansion. However, directly increasing the width or depth of dense models often yields limited benefits (see Sect.~\ref{sect: dense v.s. moe}). Inspired by Deepseek V3 \cite{liu2024deepseek}, we replaced the original DiT's FFN with MoE GLU \cite{dauphin2017language}, enabling a larger model capacity and relatively lower FLOPs requirements. In practice, within each MoE block, we use a limited greedy algorithm via a token router to allocate tokens to 2 experts out of a total of 4 different experts. A shared expert also processes all tokens. We use a load balancing loss to prevent imbalanced expert load during training. The background is provided in Suppl. Sect. {\color{iccvblue} 1.1}.
    
    \textbf{\textit{Rotary positional embedding (RoPE)}}: RoPE~\cite{su2024roformer}, a relative positional embedding method, has been shown to possess better sequence resolution and length extrapolation capabilities compared to absolute positional embeddings. Therefore, we adapted the 1D RoPE to 2D RoPE and applied it within each attention layer to enable the model's perception of position and improve its length extrapolation capabilities.
    
    \textbf{\textit{Per-block absolute positional embedding}}: The original DiT adds 2D sine absolute PE to the input latent before the backbone. We found that this can lead to image artifacts during variable resolution inference. We opt to add learnable absolute positional encodings before \textit{each} block to assist RoPE in eliminating these artifacts. Specifically, the absolute PE are generated as follows:
    \begin{equation}
    \begin{cases}
        pe_h=\text{MLP}([0,\cdots,h-1]),\\
        pe_w=\text{MLP}([0,\cdots,w-1]),\\
        pe=\text{BroadcastAdd}(pe_h, pe_w),
    \end{cases}
    \end{equation}
    where BroadcastAdd broadcasts $pe_h$, $pe_w$ the first and second dimensions and adds them together. Then we share the same $pe$ across all layers and add it with the feature.
    
    \textbf{\textit{Attention value residual}}:
    Due to the network's depth, vanishing gradients can lead to a decline in the network's learning ability, even with a considerable capacity. Inspired by \cite{zhou2024value}, we modify the value part in attention operator to:
    \begin{equation}
        V^l=(1-\eta)\cdot W^V X+ \eta V^{l-1},
    \end{equation}
    where $l$ is the index of the current layer and $V$ is the value, and $W^V$ denotes the projection weight. It means that we preserve the last layer value and do Lerp addition with the current value. $\eta$ is initiated to be 0 and set to be learnable. 
    
    \textbf{\textit{Image-based LoRA AdaLN conditioning}}:
    Since the network's condition is no longer a class label, we modified the condition to be a concatenation of the degraded latent pairs $Z_1, Z_2$ and the prompt embedding $P$ encoded by Flan-T5~\cite{chung2024scaling}.
    After then, $P$ is fed into the self-attention block to enable joint modeling. 
    As seen in Eq. \eqref{eq: ada-ln}, adaLN requires generating six different activations, which is expensive when the condition is the latent feature. We employ the LoRA to decompose the MLP into two smaller ones:
    \begin{equation}
    \alpha_1,\beta_1,\gamma_1,\alpha_2,\beta_2,\gamma_2=\text{MLP}_2(\text{MLP}_1(c)),
    \end{equation}
    where the first MLP output channel size is $d^\prime <d$, thus decreases the parameter counts.
    
    \textbf{\textit{Biased-inductive convolutions before attention}}:
    We observed that using only attention and MoE does not perform well with blur degradation. We attribute this to a lack of sufficient inductive bias. To address this, we incorporate a NAFNet block \cite{chen2022simple} before each attention block to provide a certain degree of inductive bias while incurring only a small increase in parameter counts and FLOPs. We find that this significantly improves the restoration and fusion performance for blur-related degradations.

\subsection{Regression and Flow Matching Models}


\newcommand{\best}[1]{{\cellcolor{red!10}\textbf{#1}}}
\newcommand{\second}[1]{{\cellcolor{blue!10}\textbf{#1}}}
\definecolor{green1}{RGB}{158, 200, 185}
\newcommand{\up}{$\uparrow$}
\newcommand{\down}{$\downarrow$}

\SetCommentSty{myCommentStyle}
\newcommand\myCommentStyle[1]{\textcolor{green1}{#1}}
\SetKwComment{Comment}{$\triangleright$ }{}
\SetKwComment{tcp}{$\triangleright$ }{}
\newcommand{\linecomment}[1]{\tcp*[f]{#1}}
\normalem

\begin{algorithm}[!t]
\caption{Sampling of flow matching.}
\label{algo: midpoint-sampler}
\KwIn{Degraded image pairs: $[X_0, X_1]$, prompt: $\bar P$, trained flow model $v_\theta(\cdot)$, diffusion norm $g$, sampling steps $T$, encoders $E_{img}, E_{LLM}$.}
\KwOut{Sampled clean fused image $Y$.}
$\mathbf T\gets \text{arrange}(0, 1, T)$\linecomment{Timesteps}\\
$Z_0^m, Z_1^m \gets E_{img}(X_0), E_{img}(X_1)$ \linecomment{Image latents}\\
$P\gets E_{LLM}(\bar P)$ \linecomment{Encode the prompt}\\
$Z_0\gets \mathcal N(0, \mathbf I_d)$\linecomment{Flow noise where $d=
\dim(Z_0)$}\\
\For{$t_{\tau_1}, t_{\tau_2}\gets \{\mathbf T[:-1], \mathbf T[1:]\}$}{
    $t_{mid}\gets (t_{\tau_1} +t_{\tau_2})/2$ \\
    $\Delta t\gets (t_{\tau_2}-t_{\tau_1})/2$ \\
    $Z_{t_{mid}}\gets Z_{\tau_1}+v^{(1)}_\theta(Z_t,t_{\tau_1},[Z_0^m,Z_1^m,P]) \Delta t$ \\
    \Comment{Reuse the velocity or not.}
    $v_{t_{mid}}\gets v_\theta(Z_t,t_{mid},[Z_0^m,Z_1^m,P])$ or $v^{(1)}_{\theta}$ \\
    \Comment{Stochastic flow sampling \cite{ma2024sit}.}
    $\kappa \gets g\cdot \sin^2 (\pi t_{mid})$ \linecomment{Diffusion coefficient}\\
    $\epsilon\gets \mathcal N(0, \mathbf I_d)$\linecomment{Diffusion noise}\\
    $Z_{t_{\tau_2}} \gets v_{t_{mid}} + \kappa \frac{tv_{t_{mid}}-v_{t_{mid}}}{(1-t)^2+t(1-t)} \Delta t+\sqrt{2 \kappa \Delta t}\cdot \epsilon$
}
$Y\gets D_{img}(Z_1)$ \linecomment{Decode into pixel space}
\end{algorithm}

\subsubsection{Training Losses}
Given that our model operates within a continuous latent space, we explore two distinct modeling approaches: 1) a deterministic regression model, analogous to prior works; and 2) a generative flow matching model, trained using a flow matching loss and conditioned on the degraded image latents and encoded prompts.
For the regression model $f_\theta(\cdot)$, we remove the timestep embedder, and the direct loss is set to the MSE loss with respect to the GT:
\begin{equation}
    \mathcal L_{reg}=\|f_\theta(Z^{m}_0,Z_{1}^m,P) - Z_{GT}\|_2^2,
\end{equation}
where the superscript $m$ denotes the input modality latents.
Additionally, we can incorporate an auxiliary fusion loss to further enhance network performance. Specifically, we decode the output $\tilde Z$ back into \textit{pixel space} as $\tilde X$ and then compute a conventional fusion loss \cite{ma2022swinfusion}:
\begin{equation}
\mathcal L_{aux} = \sum_{i=0}^{m-1} \|\tilde X-X_i\|_1 + \|\nabla \tilde X-\nabla X_i\|_1,
\end{equation}
where $\tilde X = D(f_\theta(Z_0^m,Z_1^m,P))$ and $\nabla(\cdot)$ denotes the gradient operator. When training the regression model, we use $\lambda$ to balance the two losses:
\begin{equation}
    \mathcal L_{total}=\mathcal L_{reg}+\lambda \mathcal L_{aux}. \label{eq: total-loss}
\end{equation}

When tackling restoration tasks with weak prior information, such as snow and rain removal, we resort to generative methods, \textit{i.e.}, flow matching. This allows us to leverage the advantages of generative modeling for improving fusion quality. Specifically, utilizing the flow matching loss in Eq. \eqref{eq: flow-loss}, our training objective is as follows:
\begin{equation}
\begin{aligned}
    \mathcal{L}_{flow}&=\mathbb{E}_{GT\in p_1, X_0\in\mathcal{N}(0,\mathbf{I})} \\ 
    &\|v_\theta(Z_t, t, Z_0^m, Z_1^m P) - (GT-X_0)\|_2^2.
\end{aligned}
\end{equation}
Notably, $\mathcal L_{aux}$ remains applicable by simply converting the predicted velocity into a predicted $\tilde Z_1$. The total loss in Eq. \eqref{eq: total-loss} can be accessed by replacing $\mathcal L_{reg}$ by $\mathcal L_{flow}$.

\begin{figure*}[!t]
    \includegraphics[width=\linewidth]{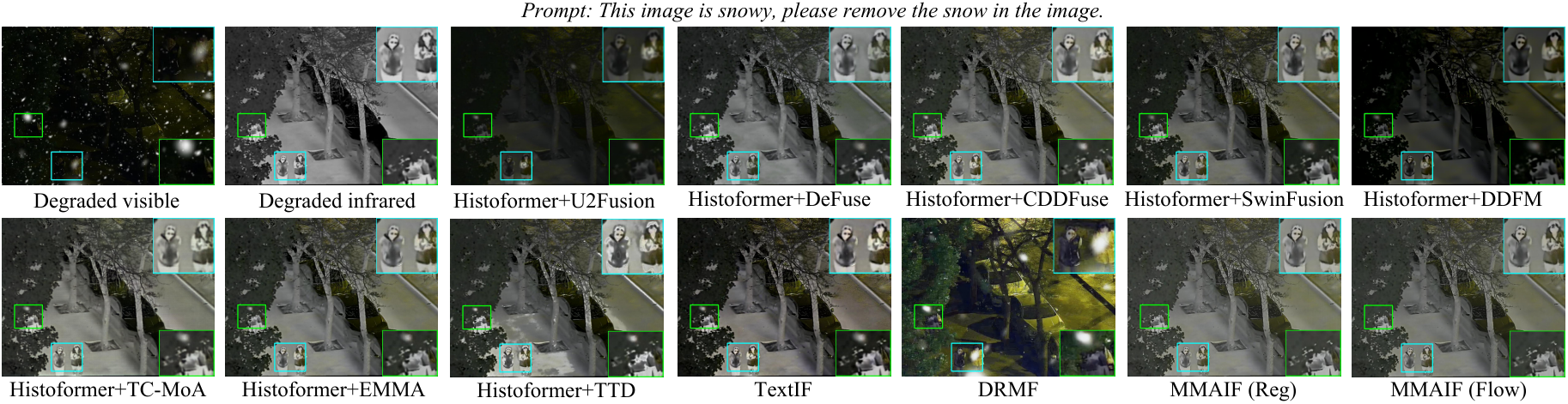}
    \includegraphics[width=\linewidth]{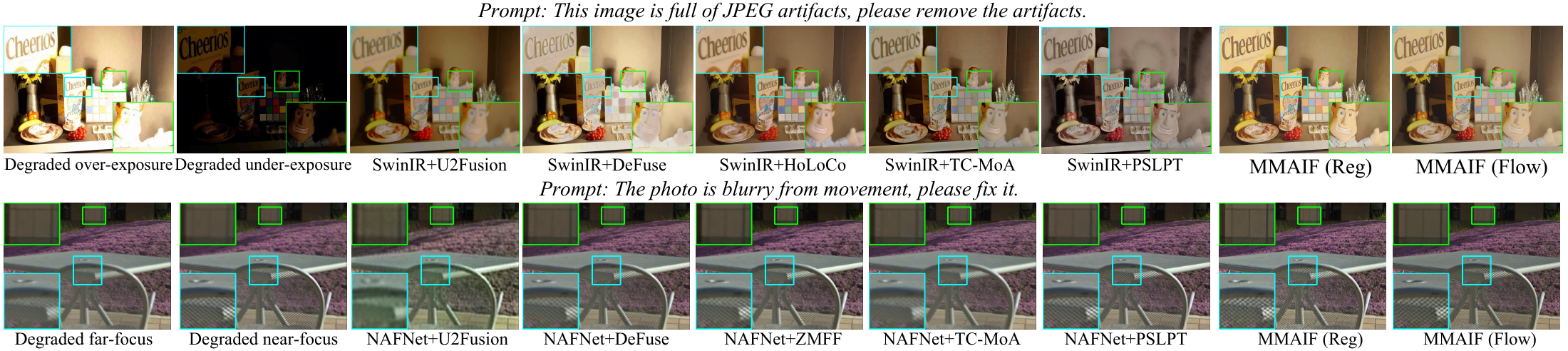}
    \caption{Qualitative comparisons of previous methods and proposed MMAIF on VIF LLVIP, MEF SICE, and MFF RealMFF datasets.}
    \label{fig: vif_mef_mff}
    \vspace{-1.5em}
\end{figure*}

\subsubsection{Fast Sampling for Flow Matching}
\label{sect: flow-matching-velocity-resuing}
During inference, the regression model directly takes pairs and prompts as input. For the flow matching model, we can turn the ODE flow path to the SDE stochastic path \cite{ma2024sit} to reduce the accumulation error.
 Using a high-order sampler (\textit{e.g.}, Dorip) does not mean a low number of function evaluations (NFEs), because one sampling step may call the function many times.
We propose a \textit{midpoint sampler reusing the predicted velocity from intermediate timesteps}, thereby accelerating from the original Euler 100 steps to Midpoint 5 steps.
The rationale behind is that if a flow model is well-trained, the flow (\textit{i.e.}, velocity) can be considered sufficiently smooth. In the midpoint ODE sampler with two NFEs in one step, when sample from $t_{\tau_1}$ to $t_{\tau_2}$, with the midpoint timestep defined as $t_{mid} = \frac{t_{\tau_1} + t_{\tau_2}}{2}$, the sampling process is formulated as:
\begin{equation}
\begin{cases}
X_{t_{mid}} = X_{t_{\tau_1}} + v^{(1)}_\theta(X_{t_{\tau_1}}, t_{\tau_1}, c) \cdot \frac{t_{\tau_2} - t_{\tau_1}}{2}, \\
X_{t_{\tau_2}} = X_{t_{mid}} + v^{(2)}_\theta(X_{t_{mid}}, t_{mid}, c) \cdot \frac{t_{\tau_2} - t_{\tau_1}}{2}.
\end{cases}
\end{equation}
We can reuse the velocity at $t_{\tau_1}$ (\textit{i.e.}, $v^{(2)}_\theta := v^{(1)}_\theta$) to replace the second SDE velocity model call. The midpoint sampling algorithm is provided in Algo.~\ref{algo: midpoint-sampler}.

These two version models imply that for image pairs with mild degradation, we can apply a regression-based model to achieve rapid fusion of degraded images; whereas for severe cases, we can leverage a flow-based model to progressively refine and obtain high-quality images.

\section{Experiments}
In this section, we first detail the data collection process for the degraded/clean/GT/prompts datasets, followed by a description of the implementation details of our model, comparisons to other works, and the metrics used. Subsequently, we present qualitative and quantitative analyses, along with ablation studies and discussions.

\subsection{Collection of Degraded Image Dataset}
We generated the degraded/clean/GT/prompts data using the degradation pipeline described in Sect.~\ref{sect: degraded-pipeline}. Our dataset relies on previous clean datasets, including three tasks, 1) VIF task: MSRS \cite{Tang2022PIAFusion}, M3FD \cite{liu2022target}, LLVIP \cite{jia2021llvip}; 2) MEF task: SICE \cite{Cai2018deep}, MEFB \cite{zhang2021benchmarking}; 3) MFF task: RealMFF \cite{zhang2020real}, WFF-WHU \cite{zhang2021mff} datasets.
In total, we synthesized 157500 pairs of data for training, 31500 pairs for validation, and 6300 pairs for testing, encompassing various image aspect ratios and varying degrees of degradation (\textit{i.e.}, $n\in \{1,2,3\}$). More details can be found in the Suppl. Sect. {\color{iccvblue} 3}.

\subsection{Implementation Details}
For the modified DiT architecture, we set the total number of layers to 16, the patch size to 2, the hidden size to 576, the LoRA dimension $d^\prime$ to 128, the number of attention heads to 12, and the RoPE base to 2000. This configuration results in a total of 282M parameters, with approximately 185M being activated during inference.
We employ the AdamW optimizer \cite{loshchilov2017decoupled} with the cautious mask \cite{liang2024cautious} to accelerate the model training. The base learning rate is set to $2e^{-4}$, and we use a cosine warmup learning rate scheduler with 1000 warmup steps. The batch size is set to 8, the total number of steps is set to 200k and we utilize gradient checkpointing during training. The gradient accumulation step is set to 3. All models are trained under the BF16 mix precision for 2 days on 2 NVIDIA RTX 4090 GPUs.

\subsection{SOTA Competitors and Metrics}
Due to the limited number of methods specifically addressing degraded image fusion, we primarily compare recent approaches such as TextIF \cite{yi2024textif} and DRMF \cite{tang2024drmf}. Additionally, we consider some restoration + fusion methods tailored to specific degradations:
1) SwinIR \cite{liang2021swinir} for denoise, super-resolution, de-JEPG; 2) NAFNet \cite{chen2022simple} for deblur; 3) PromptIR \cite{potlapalli2306promptir} for de-haze and de-rain; 4) Histoformer \cite{sun2024restoring} for snow removal.
Fusion methods include 1) U2Fusion~\cite{xu2020u2fusion}, SwinFusion~\cite{ma2022swinfusion}, DDFM~\cite{zhao2023ddfm}, EMMA~\cite{zhao2024equivariant}, and TTD \cite{cao2024ttd} for VIF task; 2) U2Fusion \cite{xu2020u2fusion}, DeFuse \cite{Liang2022ECCV}, HoloCo \cite{liu2023holoco}, TC-MoA \cite{zhu2024task}, and PSLPT \cite{wang2024general} for MEF task; and 3) previous MEF task methods and additionally involved ZMFF \cite{hu2023zmff} for MFF task. As for metrics, we employ VIF, $Q^{A/BF}$ \cite{ma2019infrared}, BRISQUE \cite{mittal2012no}, MUSIQ \cite{ke2021musiq}, and CLIPIQA \cite{wang2023exploring} metrics for the VIF task. For MEF and MFF tasks, we replace the VIF metric with $Q_{cv}$ \cite{zhang2021benchmarking} metric.

\subsection{Results of Different Tasks}
The results of the three image fusion tasks, VIF, MEF, and MFF, are presented in Tabs.~\ref{tab: vif} and~\ref{tab: mef_mff}, with some qualitative comparisons shown in Fig.~\ref{fig: vif_mef_mff}. It can be observed that our MMAIF outperforms previous restoration+fusion methods as well as recent all-in-one methods (\textit{i.e.}, TextIF and DRMF) in almost all metrics. Additionally, flow-matching-based models perform better than the regression-based model; however, under mild degradation conditions, the regression-based model also performs sufficiently well. More comparisons and visual results are provided in Suppl. Sect. {\color{iccvblue} 5}.

\begin{table*}[t]
\centering
\setlength{\tabcolsep}{2pt}
\caption{Quantitative metrics of MSRS, M3FD, and LLVIP tasks in degraded VIF task. ERN denotes an existing restoration network. Reg and FM mean regression and flow matching, respectively. The best and second-best results are colored in \best{red} and \second{blue}.}
\vspace{-0.5em}
\label{tab: vif}
\resizebox{\textwidth}{!}{%
\begin{tabular}{c|ccccc|ccccc|ccccc}
\toprule
\hline
\multirow{2}{*}{Methods} & \multicolumn{5}{c|}{\textbf{LLVIP Dataset}} & 
\multicolumn{5}{c|}{\textbf{M3FD Dataset}} & \multicolumn{5}{c}{\textbf{MSRS Dataset}} \\
 & VIF & $Q^{A/BF}$ & BRISQUE & MUSIQ & CLIPIQA & VIF & $Q^{A/BF}$ & BRISQUE & MUSIQ & CLIPIQA & VIF & $Q^{A/BF}$ & BRISQUE & MUSIQ & CLIPIQA \\
 \hline
ERN+U2Fusion \cite{xu2020u2fusion} & 0.429 &	0.081 &	76.023 &	29.483 &	0.171 
 & 0.462&	0.117&	65.217&	29.144&	0.184
 &  \best{0.616} &	0.101& 	81.787 &	19.539& 	0.110 
 \\ 
ERN+DeFuse \cite{Liang2022ECCV} &  0.552 &	0.272 	&35.479 	&45.528 &	0.284 
& 0.424 &	0.236& 	39.637 &	48.227 &	0.269  &0.563 	& 0.304 &	43.676 &	32.496 &	0.154 
\\
ERN+SwinFusion \cite{ma2022swinfusion} & \second{0.571} &	0.289 &	30.122 &	47.341 & 	0.245 
 & 0.459&	0.273& \second{33.506}	&51.574	& 0.230
 & 0.517 &	0.296 &	38.141 	&34.995 &	0.167 
 \\
ERN+CDDFuse \cite{Zhao_2023_CVPR}&  0.527 & 0.359 &	28.877 &	49.785 &	0.257 
 & 0.415 &	0.362 &	38.761 &	54.301&	0.254 
 &0.465&	0.358 &	35.898&	\second{37.911}& 	0.172 
 \\
ERN+DDFM \cite{zhao2023ddfm}& 0.515 &	0.189 &	46.041 &	45.913& 	0.267 
 &0.439	&0.193	&38.840&	49.398	&0.238
 & 0.472 &	0.151 	&56.606 	&31.988 &	0.152 
  \\
ERN+TC-MoA \cite{zhu2024task} & 0.486 &	0.237 &	42.774 &	38.295& 	0.222 
 &  0.385 &	0.193 	&45.964 	&36.160 &	0.270 
  & 0.484 &	0.267 &	50.518 &	24.775& 	0.164 
\\
 ERN+TTD \cite{cao2024ttd} & 0.457 & 0.342 & 32.032 & 48.137 & 0.310 &
0.358 & 0.352 & 37.445 & 53.577 & 0.270 & 0.456 & 0.351 & 41.484 & 37.467 & 0.170 
\\
ERN+EMMA \cite{zhao2024equivariant}& 0.479 &	0.368 &	\second{27.788} &	49.277 &	0.242 
 & 0.387 &	0.372 &	37.455& 	52.923 &	0.260 
&0.472 &	0.354 &	\second{32.165} 	&36.306 	&0.164 
   \\
\hline
TextIF \cite{yi2024textif} & 0.456 &	\best{0.394} &	28.161 	&49.499 &	0.271
 &  0.409 &	0.397 	& 33.627 &	\second{54.307} &	0.271 
& 0.480 &	0.362 &	36.731 	&37.358 	&0.180 
 \\
DRMF \cite{tang2024drmf} &  0.339 &	0.218 &	41.685 &	39.520 &	0.241  
  &  \best{0.506} &	\second{0.379} &	42.403	& 48.898 &	0.256 & 0.538&	0.390 &42.012 &	34.713 &	0.132

 \\
\hline
Ours (Reg) & 0.556 &	0.374 & 29.351  &	\second{50.906}  & 	\second{0.312}
 &0.427 	& \best{0.381} &	\best{32.860} & 	53.503	& \second{0.271}
 & 0.530 &	\second{0.398} & 35.328 &  34.589  &		\best{0.203}
\\
Ours (FM) & \best{0.604} &	\second{0.384} 	& 	\best{27.535}	& \best{53.277} 	& \best{0.330}
 & \second{0.464} &	\best{0.381} &	35.696 	& \best{55.423}  &	\best{0.274} 
 & \second{0.586} &	\best{0.406} 	& \best{32.076} &		\best{40.413} 
& \second{0.217}
\\
\hline
\bottomrule
\end{tabular}%
}
\vspace{-1em}
\end{table*}

\begin{table*}[t]
\caption{Quantitative metrics of SICE and RealMFF task in degraded MEF and MFF tasks. ERN denotes the existing restoration network.  Reg and FM mean regression and flow matching, respectively. The best and second-best results are colored in \sethlcolor{red!10}\hl{red} and \sethlcolor{blue!10}\hl{blue}.}
\vspace{-0.5em}
\label{tab: mef_mff}
\resizebox{\linewidth}{!}{
\begin{tabular}{c|ccccc|c|ccccc}
\toprule
\hline
\multirow{2}{*}{Methods} & \multicolumn{5}{c|}{\textbf{SICE MEF Dataset}} & \multirow{2}{*}{Methods} & \multicolumn{5}{c}{\textbf{RealMFF MFF Dataset}}  \\
& $Q_{cv}$ & $Q^{A/BF}$ & BRISQUE & MUSIQ & CLIPIQA &  & $Q_{cv}$ & $Q^{A/BF}$ & BRISQUE & MUSIQ & CLIPIQA  \\
\hline
ERN+U2Fusion \cite{xu2020u2fusion}& 397.979 &	0.108 	&66.115 &	29.594 &	0.211 &
ERN+U2Fusion\cite{xu2020u2fusion} & 168.027	&0.471 	&45.982 	&39.606 &	0.224 \\
ERN+DeFuse \cite{Liang2022ECCV} & 364.307 &	0.324& 	31.793 &	52.109& 	0.357 &
 ERN+DeFuse \cite{Liang2022ECCV} & 201.942 & 0.465 & 46.628 & 37.817 & 0.197 \\
ERN+HoLoCo \cite{liu2023holoco} & 403.183 &	0.251 &	34.222 &	54.152 &	0.360 &
 ERN+ZMFF \cite{hu2023zmff} & 315.978 	&0.469 	&50.398 &	42.405& 	0.250 
 \\ 
ERN+TC-MoA \cite{zhu2024task}& 389.607 &	0.235 &	42.859 &	40.769 &	0.312 &
 ERN+TC-MoA \cite{zhu2024task} &438.497 &	0.268 &	57.361 	&20.969 &	0.137 
 \\
ERN+PSLPT \cite{wang2024general} & 395.880 &	\best{0.363} &	27.449 & 	58.593& 0.326 &
 ERN+PSLPT \cite{wang2024general} & 201.088 & 0.490 & 40.938 & 45.260 & 0.238 \\
\hline
Ours (Reg)  & \second{279.903} & \second{0.343}	& \second{21.239} &	 \second{63.600} & \second{0.401} &
 Ours (Reg) & \second{106.687} 	& \second{0.492} &	\best{35.371} &	 \second{50.614} & 	\second{0.290}
 \\
Ours (FM) & \best{268.068}	 & 0.336	& \best{12.825} 	& \best{68.069}	& \best{0.410} & 
 Ours (FM) & \best{105.973} &	\best{0.503} &	\second{37.069} &	\best{52.768} &	\best{0.292}
 \\  
\hline
\bottomrule
\end{tabular}
}
\vspace{-1.4em}
\end{table*}

\subsection{Combinations of Degradations}
\begin{figure}
    \centering
    \includegraphics[width=\linewidth]{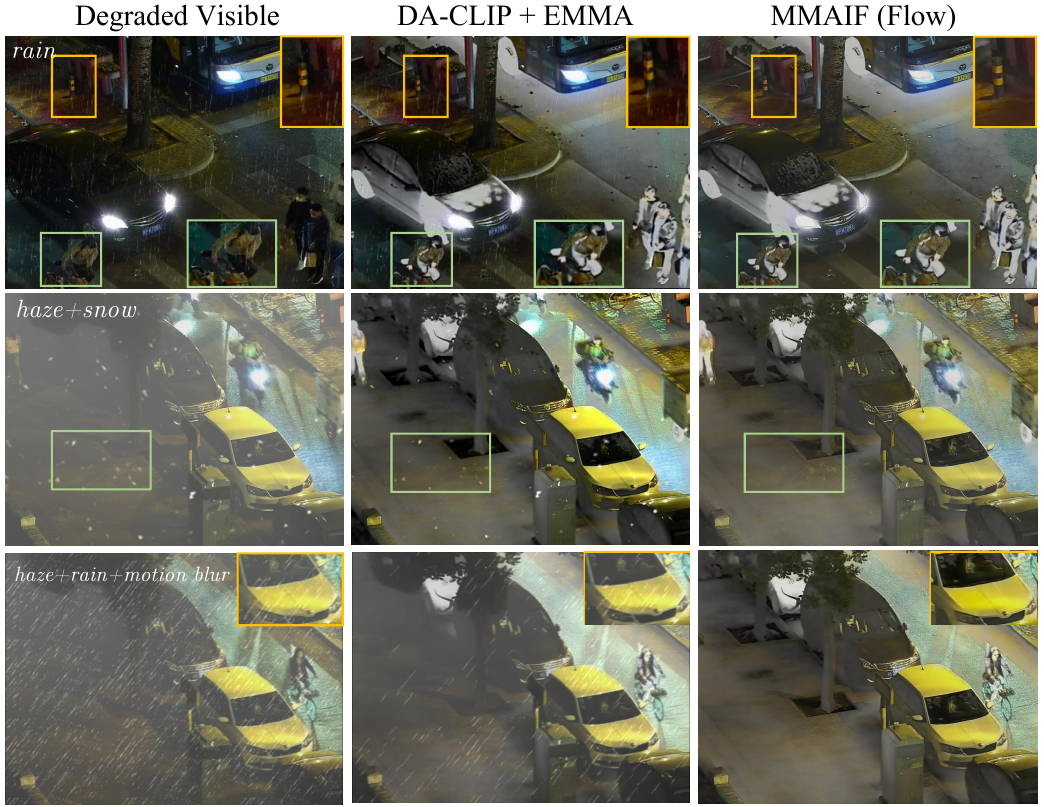}
    \caption{Combined degradations and fused results.}
    \label{fig: degs1-3}
    \vspace{-1.5em}
\end{figure}

\begin{table}
\caption{Results of total $n$ degradations in LLVIP dataset. ``R+F'' denotes degradation-level all-in-one DA-CLIP + EMMA pipeline.}
\label{tab: multi-degs_comps}
\resizebox{\linewidth}{!}{
\begin{tabular}{l|ccccc}
\toprule
\multirow{2}{*}{Config} & \multicolumn{5}{c}{\textbf{LLVIP Dataset}} \\ 
& VIF & $Q^{A/BF}$ & BRISQUE & MUSIQ & CLIPIQA \\
\hline
$n=1$ R+F & 0.496 & 0.370 & 28.647 & 52.015 & 0.286 \\
$n=1$ Ours & \best{0.604} &	\best{0.384} 	& 	\best{27.535}	& \best{53.277} 	& \best{0.330} \\
\hline
$n=2$ R+F & 0.450 & 0.346 & 33.256 & 48.589 & 0.264 \\
$n=2$ Ours & \best{0.601} & \best{0.380} & \best{28.346} & \best{51.807} & \best{0.326} \\
\hline
$n=3$ R+F & 0.431 & 0.325 & 35.761 & 44.285 & 0.247 \\
$n=3$ Ours & \best{0.584} & \best{0.366} & \best{30.432} & \best{48.691} & \best{0.317} \\
\bottomrule
\end{tabular}
}
\vspace{-1em}
\end{table}

Unlike previous all-in-one methods, our framework supports not only multiple fusion tasks but also various types of degradation. We choose recent all-in-one DA-CLIP \cite{luo2023controlling} to act as the restoration model. Tab.~\ref{tab: multi-degs_comps} and Fig.~\ref{fig: degs1-3} demonstrate the outstanding fusion performance of our method under scenarios where two or three degradations coexist. 

\subsection{Ablations and Discussion}
We conducted analyses and discussions on the LLVIP dataset to verify the effectiveness of our proposed modules.

\subsubsection{Dense Model \textbf{\textit{v.s.}} MoE Model}
\label{sect: dense v.s. moe}
To demonstrate the effectiveness of MoE, we compare it with a dense DiT and a dense U-net with similar number of parameters. The results of the VIF task are shown in Tab.~\ref{tab: ablations} (a-b), indicating that despite having a similar number of parameters, the MoE version of DiT still outperforms the dense models. This shares a similar observation with \cite{liu2024deepseek}.

\subsubsection{Improved DiT Module Ablations}
For each of the proposed improvements in Sect.~\ref{sect: modernized-dit}, we conducted ablation studies module by module, with the results presented in Tab.~\ref{tab: ablations} (c-f). It can be observed that MoE yields significant performance gains, RoPE and per-block PE further enhance performance, attention value residual slightly improves the outcome, LoRA and AdaLN maintain performance almost unchanged while reducing parameter size, and biased-inductive convolution also boosts fusion performance.

\begin{table}[t]
    \centering
    \caption{Whole pipeline latency under $1024\times1024$ resolution.}
    \label{tab: latency}
    \setlength{\tabcolsep}{2pt}
    \resizebox{\linewidth}{!}{
    \begin{tabular}{c|cccc}
    \toprule
        Latency & SwinIR+EMMA & DA-CLIP+EMMA & NAFNet+PSLPT & MMAIF (Reg) \\
        \hline
        Sec./Img. & 5.78+0.02 & 19.45+0.02 & 0.345+0.013 & \best{0.108} \\
    \bottomrule
    \end{tabular}
    }
    \vspace{-1em}
\end{table}

\begin{table}[t]
\setlength{\tabcolsep}{2pt}
\caption{Ablation study of dense variants, proposed modernized DiT modules, flow matching samplers, and prior $p_0$ choices.}
\label{tab: ablations}
\resizebox{\linewidth}{!}{
\begin{tabular}{l|ccccc}
\toprule
\multirow{2}{*}{Config} & \multicolumn{5}{c}{\textbf{LLVIP Dataset}} \\ 
& VIF & $Q^{A/BF}$ & BRISQUE & MUSIQ & CLIPIQA \\
\hline
(a) Dense U-net & 0.423 & 0.326 & 32.876 & 47.869 & 0.281 \\
(b) Dense DiT & 0.436 & 0.331 & 31.909 & 48.781 & 0.285 \\
\hline
(c) + MoE    & 0.582 & 0.373 & 29.805 & 52.816 & 0.313 \\
(d) + RoPE, per-blk PE &  0.593 & 0.378 & 28.672 & 52.781 & 0.314 \\
(e) + Value-residual Attn. & 0.599 & 0.380 & 28.031 & 53.209 & 0.325 \\
(g) + NAFBlock  &  0.607 & 0.388 & 27.543 & 53.730 & 0.331 \\
(f) + LoRA-AdaLN &  0.604 &	0.384 	& 27.535	& 53.277 	& 0.330\\
\hline
(h) Euler (25 steps) & 0.592 & 0.378 & 28.186 & 52.766 & 0.325 \\
(i) Euler (100 steps) & 0.605 & 0.381 & 27.582 & 53.188 & 0.331 \\
\hline
(j) $p_0=Z_0$  &  0.435 & 0.334 & 31.267 & 48.169 & 0.290  \\
(k) $p_0=Z_1$  & 0.422 & 0.320 & 32.964 & 47.558 & 0.283  \\
(l) $p_0=mean$ & 0.438 & 0.337 & 31.107 & 49.014 & 0.290\\
(m) I$^2$SB (100 steps) \cite{liu20232} & 0.601 & 0.381 & 27.549 & 53.177 & 0.328 \\
\hline
Ours (default) & 0.604 &	0.384 	& 27.535	& 53.277 	& 0.330 \\
\bottomrule
\end{tabular}
}
\vspace{-1.2em}
\end{table}

\subsubsection{Non-Gaussian Prior in Flow Matching}
We explored setting $p_0$ to $Z_0$, $Z_1$, and $\text{mean}(Z_0, Z_1)$. The results are shown in Tab.~\ref{tab: ablations} (j-l). We found that the sampled images from the trained flow-matching models contained artifacts or had poor restoration quality. We attribute this issue to the accumulation of errors due to a lack of stochasticity during sampling.  
To validate this hypothesis, we use I$^2$SB \cite{liu20232} (see Tab.~\ref{tab: ablations} (m) and Suppl. Sect. {\color{iccvblue} 1.2}) with $p_0=mean$ to obtain a stochastic sampling path (background is in Suppl. Sect. {\color{iccvblue} 3}). This variant performs well, confirming our conjecture.

\subsubsection{Inference Latency}
Our framework operates in the latent space, significantly reducing the overhead of DiT. Compared to methods operating in pixel space, our approach offers substantial inference advantages, as demonstrated by the latency comparisons with previous methods provided in Tab.~\ref{tab: latency}. Our MMAIF can restore and fuse a $1024\times 1024$ resolution image during \textit{only 0.1 seconds}, far less than previous restoration+fusion pipelines, and MMAIF (flow) only needs \textit{less than 0.6 seconds} due to the velocity reusing technique proposed in Sect.~\ref{sect: flow-matching-velocity-resuing}. This shows that working in latent space can bring much lower inference latency.

\section{Conclusion}
We propose a unified framework that addresses multiple fusion tasks and handles various image degradations through restoration and fusion. Key components include a simulated real-world degradation pipeline, a DiT model tailored for image fusion, and regression/flow matching models operating in the latent space. By leveraging language guidance, the framework achieves targeted restoration with efficient inference. Extensive experiments validate its superior performance and broad applicability.

\normalem
{
    \small
    \bibliographystyle{ieeenat_fullname}
    \bibliography{main}
}

\end{document}